\documentclass[letterpaper]{article} 
\usepackage[]{aaai24}  
\usepackage{times}  
\usepackage{helvet}  
\usepackage{courier}  
\usepackage[hyphens]{url}  
\usepackage{multirow}
\usepackage{graphicx} 
\usepackage{natbib}

\usepackage{tabularray}

\usepackage{caption} 
\usepackage{algorithm}
\usepackage{array}
\newcolumntype{P}[1]{>{\centering\arraybackslash}p{#1}}
\usepackage{tabularx}
\usepackage{algorithmic}

\usepackage{newfloat}
\usepackage{listings}
\usepackage{multirow}
\usepackage{amssymb}
\DeclareCaptionStyle{ruled}{labelfont=normalfont,labelsep=colon,strut=off} 
\floatstyle{ruled}
\newfloat{listing}{tb}{lst}{}
\floatname{listing}{Listing}

\usepackage{booktabs}

\newcommand{\scl}[2][c]{%
  \begin{tabular}[#1]{@{}c@{}}#2\end{tabular}}
\title{What Is Required for Empathic AI? It Depends, and Why That Matters for AI Developers and Users}
\author{
    Jana Schaich Borg\textsuperscript{\rm 1} and Hannah Read\textsuperscript{\rm 2}
    }

\affiliations {
    \textsuperscript{\rm 1}Social Science Research Institute, Duke Institute for Brain Sciences, Duke University\\
    \textsuperscript{\rm 2}Kenan Institute for Ethics, Duke University\\
    js524@duke.edu, hannah.read@duke.edu
}

\begin{document}
\maketitle
\begin{abstract}
Interest is growing in artificial empathy, but so is confusion about what artificial empathy is or needs to be. This confusion makes it challenging to navigate the technical and ethical issues that accompany empathic AI development. Here, we
outline a framework for thinking about empathic AI based on the premise that different constellations of capabilities asso-
ciated with empathy are important for different empathic AI applications. We describe distinctions of capabilities that we
argue belong under the empathy umbrella, and show how three medical empathic AI use cases require different sets of these capabilities. We conclude by discussing why appreciation of the diverse capabilities under the empathy umbrella is important for both AI creators and users.

\end{abstract}

\raggedbottom

\section{Introduction}

According to one author, ``In order to achieve artificial general intelligence, an AI must use empathy to make decisions'' \citep{Forbes}.  Indeed, interest in empathic AI systems is growing, as indicated by recent publication titles such as ``Artificial empathy: the upgrade AI needs to speak to consumers'' \citep{bhansali2022artificial} and ``Why We Need Empathy In AI'' \citep{Washington}. Some have even suggested empathic AI may offer benefits over human empathy, such as being less resource-limited and potentially less biased towards ingroups \citep{inzlicht2023praise}.  This interest, in turn, raises the fundamental question: what capabilities does an AI system need to have to be considered empathic?  Our goal here is to provide some guidance for how to think about that question.  

Many different kinds of AI systems have been called ``empathic''.\footnote{We use quotation marks here in recognition of the lack of consensus of what empathy is, but we will drop the quotations in the rest of the manuscript for readability.}  Audi’s ``empathic mobility partner'', AI:Me, personalizes the autonomous driving experience for passengers and signals to others on the road what the car ``intends'' to do.  HumeAI offers ``The empathic AI toolkit for researchers and developers'', including tools for measuring emotions from sound or videos.  Siena, which offers autonomous chat-based customer service agents, advertises that it provides ``human empathy in every interaction''.  Researchers, too, have to navigate a dizzying number of conceptions of empathy \citep{hall2022empathy}. As one author expressed, ''there are perhaps as many definitions as there are authors in the field'' \citep{cuff2016empathy}.  Yet, despite this variety, researchers frequently have very strong and conflicting opinions about which empathy definition is the ``true'' or ''right'' one, sometimes motivating them to respond to empathic AI efforts with some version of ``well, that’s not real empathy'' \citep{mcstay2023replika, coplan2011will}.  The result is general confusion on the part of both AI creators and AI users.

Here, we outline a framework for thinking about empathic AI that we believe will facilitate development of useful empathic systems, make it easier for interdisciplinary researchers to test and audit them, and support practices that minimize ethical harms that occur when the nature of an empathic system is misunderstood.  Rather than focus on what ``true'' empathy is, we argue that different constellations of capabilities associated with empathy are important for different applications, and AI creators must think carefully about---and perhaps collect evidence about---which of these capabilities will be necessary, helpful, or even counterproductive for the specific empathic AI use cases they are working on.  

To introduce our framework and motivate our recommendations, we begin by discussing three empathic AI use cases in medicine,  a domain where enthusiasm for empathic AI has been growing.  Next, we describe distinct capabilities that have been recognized under the larger empathy umbrella, draw attention to ways the umbrella is both wider and more fine-grained than previous review papers have emphasized, and use those distinctions to illustrate how the empathic abilities needed for each medical empathic AI use case likely differ.  Finally, we discuss why appreciating these ``fine cuts'' of empathy is important for both AI creators and users.  

\section{Three Empathic AI Use Cases in Medicine}

\emph{Medical question answerers.} The first empathic AI application we consider is AI that responds to medical questions virtually.  People often use forums like iCliniq, Lybrate, FindaTopDoc, JustAnswer, and others to ask doctors medical questions without having to schedule in-person appointments.  When doctors provide answers too abruptly or dispassionately on these forums, users sometimes become too fearful or ashamed to seek additional medical help \citep{Mok}.  Answers conveyed with empathy, on the other hand, help patients feel comfortable continuing to ask questions, and may make it more likely that patients ultimately follow the medical advice they find in the forums.  The goal of empathic medical question answerer AIs is to give answers to patients in a way that maximizes patients’ likelihood of being receptive to the information presented, following relevant advice, and being willing to ask questions again in the future. 
\linebreak 

\emph{AI care assistants.} The second application is an assistant to the elderly, cognitively impaired, or disabled.  These AI care assistants are supposed to help patients with things they need to do, provide reminders, give reassurance and encouragement, help patients navigate dangerous situations, and perhaps provide some sorts of company, entertainment, or mental stimulation.  For example, ElliQ is a robot that has been used by the state of New York to remind seniors to take their medications, provide wellness suggestions, and provide ``proactive and empathetic'' conversations \citep{Muoio}.  Users will most likely not use these kinds of AI products if doing so makes them feel worse than they already do or if the AIs are not tailored to the specific issues the users need help with.  In fact, using AIs in such cases could actually be dangerous and lead users to engage in physical harm to themselves or others (as has been alleged in the case of one Belgian man who committed suicide after chatting with an AI chatbot; \citealt{Vice}) or ignore critical safety advice.  Even AIs that listen in an emotionally neutral way can make users feel insignificant \citep{halpern2001detached}.  So the role of empathy in AI care assistants, as we are considering them here, is to convey sensitivity to users’ unique needs, beliefs, and vulnerabilities, and allow the AIs to interact with users in a way that feels sufficiently comfortable to the users, given the vulnerable situation many users will be in. 
\linebreak 

\emph{AI care providers.} The third application is an entity that is meant to provide care for patients.  This application is different than the AI care assistant, because in this case, it is deemed essential that the patient feels truly cared for, not just aided or assisted.  Mental health therapy and long-term treatment for chronic illness are two applications where feeling cared for correlates with better health outcomes and is believed to be critical \citep{vitinius2018perceived}.  In such settings, it has been argued that patients must perceive that the care provider engages with, attends to, understands, and respects their experience, and that the care provider is genuinely motivated to help them because they feel and think the patient’s well-being has intrinsic value and that the patient deserves dignity \citep{montemayor2022principle, perry2013understanding, portacolone2020ethical}.  Some argue a patient must sense that the caretaker has real and appropriate feelings in response to the patient’s joy and suffering to achieve the documented medical benefits.  Simply being able to label and understand the patient’s feelings is not enough because care involves a ``sharing of emotional feeling and connecting with the patient at an emotional level'' \citep{jeffrey2016empathy}.  The goal of empathic AI care providers is to manifest the same types of unique positive health impacts human medical providers that are perceived to ``care about their patient'' have, without manipulating, misleading, or disrespecting patients.  Even if it is reasonable to question whether patients truly need to feel cared for to achieve those desired medical benefits, for our purposes here,  AI care provider creators do want patients to feel their AI exhibits the qualities described above.

\subsection{Something is Different about these Empathic AIs, but What?}

We posit that the aforementioned three empathic AI use cases require different constellations of abilities.  Even so, AIs functioning successfully in these use cases would be displaying empathy by at least one of the plethora of definitions proposed by reputable researchers \citep{hall2019empathy}.  There are valid research questions related to whether a specific AI system meets a chosen theoretical definition of what human empathy is believed to be.  For many interested in building empathic AI systems, though, the more important task is to clarify what constellation of specific abilities from within the empathy umbrella the empathic system one wants to build needs to have.    In the next section, we offer distinctions drawn from a wide variety of disciplines that are useful for thinking about these empathic abilities.  In all discussions, we use ``empathizer'' to refer to the agent who is supposed to have empathy and ``target'' to refer to the agent the empathizer has empathy for.

\section{``Fine cuts'' of Empathy: Capabilities and Distinctions under the Empathy Umbrella}
To explain important distinctions or ``fine cuts'' \citep{blair2008fine} of empathic phenomena efficiently, it is helpful to draw upon some ``broader cut'' concepts and terminology.  Most (but not all) accounts of empathy differentiate between three related, but separate, phenomena: cognitive empathy (or understanding of another’s experience, preferences, or knowledge), affective empathy (feelings or emotional experiences linked to another’s experience, preferences, or knowledge), and motivated empathy (other-oriented concern congruent with another’s perceived welfare, sometimes called compassion or sympathy).  Some of the fodder for distinguishing cognitive empathy from affective empathy comes from medicine, where it has been shown that people can have deficits in understanding what someone knows or believes while maintaining appropriate emotional responses to someone else’s emotions, and visa versa \citep{fletcher2020autism, winter2017social}.  In addition, neuroscience studies have shown that different brain networks are involved in cognitive empathy and affective empathy, supporting the idea that these phenomena are distinguishable and can be impacted independently \citep{stietz2019dissociating}.  Compassion and/or sympathy---which are also contested terms, but typically refer to motivations to alleviate others’ suffering---are sometimes (but not always) separated out from cognitive and affective empathy, because the thoughts and emotions we have in response to somebody else’s situation don’t always lead us to help that person or want to do so.  

Another concept incorporated into many accounts of empathy is mirroring.  The general idea is that our brains and bodies might need to have a representation of another person’s knowledge, thoughts, preferences, or feelings that is identical to what we ourselves would think, perceive, or feel in that same situation in order to understand, feel, or respond acceptably to what that other person is going through.  Some philosophical and psychological accounts of empathy postulated the presence of mirroring mechanisms long ago \citep{zahavi2010empathy}, but enthusiasm for these accounts seemed to take off more notably when evidence emerged that so-called ``mirror neurons'' in a monkey’s motor system fire both when monkeys see another monkey reach and when monkeys reach themselves \citep{iacoboni2009imitation}.  Perhaps relatedly, activity increases in overlapping brain regions when we observe, perceive, or believe someone else is having a feeling and have those feelings ourselves, especially when the feelings involve physical pain \citep{singer2009differential}.  Some accounts (like the perception-action model of empathy) argue that mirroring is at the core of all empathic phenomena \citep{preston2002empathy}.  Other accounts just acknowledge that mirroring is either one of the capabilities that belongs in the empathy umbrella, or a mechanism that contributes to capabilities under the empathy umbrella (e.g. \citet{yamamoto2017primate}). 

The notions of cognitive empathy, affective empathy, motivated empathy, and mirroring are prevalent in most empathy accounts, as noted by syntheses of the empathy literature \citep{hall2019empathy, eklund2021toward}.  However, additional distinctions and requirements emerge from the plethora of empathy accounts as well, even though some are acknowledged much less consistently across disciplines.  These less-discussed considerations are important to AI creators thinking about what their AI systems must achieve.  We will discuss these ``fine cuts'' of empathy next (see also Table 1).\linebreak

\emph{Information types and how they need to be known.} The first way of distinguishing empathic capabilities is through the type of information the empathizer is supposed to glean about the target.  The empathizer might need to know the target’s perceptions (row 1 in Table 1), cognitive phenomena (which include beliefs, thoughts, and knowledge; row 2), feelings (also referred to as emotions for our purposes; row 3), or some combination of all three.  The empathizer may be expected, or even required, to glean this information through specific combinations of their own perceptions, cognitive phenomena, or feelings (rows 4, 5, 6).  

Consider, for example, the Sally-Ann test of cognitive empathy, sometimes referred to as ``theory of mind'' \citep{baron1985does} that has been incorporated into recent benchmark tests for large language models \citep{le2019revisiting} (note that ``theory of mind'' is typically considered only one of the functions that contributes to cognitive empathy; \citealt{perry2013understanding}).  In this test, cartoon sequences illustrate Sally putting a marble in a basket and leaving the room. While she is gone, Anne removes the marble from the basket and puts it in a box. Sally comes back into the room, and a potential empathizer is asked where Sally will look for her marble.  To arrive at the correct answer of ``in the basket'', an empathizer must know that Sally did not see (or visually perceive) that the marble was moved and that, as a consequence, Sally will believe the marble is still in the basket.  The empathizer doesn’t need to know anything about what Sally feels to pass the task.  Thus, if there was a column for the Sally-Ann test in Table 1, rows 1 and 2 would be checked, but row 3 would be blank.  Since the empathizer must be able to see the Sally-Ann cartoons or hear their descriptions to pass the test and must form appropriate beliefs based on what they perceive, rows 4 and 5 would be checked as well.  However, the empathizer does not need to have any specific feelings to pass the Sally-Ann test.  

By contrast, consider ``empathic accuracy'' tests that many neuroscience and psychology studies use \citep{hall2007sources}.  In these tasks, empathizers are shown pictures of people expressing different facial expressions and are asked which emotion the person in each picture is feeling.  To pass the tests, empathizers need to know what the target feels, but not what they think, know, believe, or perceive.  Empathizers must perceive the pictures and form appropriate beliefs about them to pass empathic accuracy tasks, but do not necessarily need to feel anything specific to pass them.  So if there was a column for empathic accuracy tests in Table 1, rows 3, 4, and 5 would be checked, but rows 1, 2, and 6 would be blank.  That said, some definitions and theories of empathy state or hypothesize that empathizers come to know what a target is experiencing through experiencing similar feelings of their own, and the activation of the empathizer’s own feelings is what allows empathy to function efficiently \citep{de2017mammalian}.  Under such a conception, row 6 in Table 1 would be checked as well.\linebreak

\emph{Consciousness.} Some conceptions of empathy require that the empathizer either be ``consciously attentive'' to certain types of information or have a certain kind of conscious experience (row 7 in Figure 1).  For example, \citet{yu2018dual} say that cognitive empathy involves ``a slow and complex process with efforts, consciousness, and elaborated neural profiles'' \citep{yu2018dual}.  \citet{smith2017empathy} says ``A empathizes with B if and only if (1) A is consciously aware that B is $\phi$, (2) A is consciously aware of what being $\phi$ feels like, (3) On the basis of (1) and (2), A is consciously aware of how B feels'' \citep{smith2017empathy}. \citet{montemayor2022principle} state that some components of empathy are unconscious, some are conscious, but successful empathic communication that conveys genuine care requires ``consciously practic[ing] empathic attention'' \citep{montemayor2022principle}.  So even if not all aspects of empathy require consciousness, consciousness may be required to realize some of empathy’s benefits.

Other accounts explicitly state that empathy or some of its components can be unconscious.  This is often a stipulation for ``emotional contagion'', or ``[t]he tendency to automatically mimic and synchronize expressions, vocalizations, postures, and movements with those of another person’s, and, consequently, to converge emotionally'' \citep{hatfield1993emotional}, in particular.  Some theories, especially those from the neuroscience field, postulate that emotional contagion is the core of all other empathic processes, and at minimum a process that should be considered part of empathy \citep{preston2002empathy}.  Emphasizing that assumption, neuroscience studies of empathy often focus on automatic, unconscious brain responses while passively viewing pictures of others in pain or distress \citep{singer2009differential}.\linebreak

\emph{Accuracy.} Some conceptions of empathy require that the empathizer not only perceive, believe, and/or emotionally respond to a target’s perceptions, beliefs, or feelings, but that they do so with sufficient accuracy (row 8).  For example, it may not be enough for an empathizer to try to understand a target’s thoughts to have cognitive empathy \citep{rogers1975empathic}.  Instead, empathizers may need to make correct theories and inferences about their target’s mental states and behavior, especially if they want their targets to feel understood \citep{rogers1975empathic, spaulding2017cognitive}. 
That said, accuracy may not always be required for empathy to have an impact.  If an empathizer believes a target is sad, resonates with that imagined or perceived sadness, and takes actions to try to relieve the target of that believed sadness, many theoretical and folk accounts of empathy would still consider that evidence of affective empathy or motivated empathic concern, even if the target turned out not to feel sad after all \citep{ickes2001motivational}.  Evidence of empathic concern, even if misguided, can be sufficient for some purposes.
\\

\emph{Self-other differentiation.}  Another characteristic of many empathy definitions is that the empathizer must (or must not) have particular forms of appreciation for the differences between themselves and the target, including differences in what they are thinking, feeling, and why (row 12).  For example, \citet{singer2009differential} specify that empathy involves ``a distinction between oneself and others and an awareness that one is vicariously feeling with someone but that this is not one’s own emotion'' \citep{singer2009differential}, and \citet{decety2006human} specify that empathy is ``the ability to experience and understand what others feel without confusion between oneself and others'' \citep{decety2006human}.  

Even so, emotional contagion---one of the aforementioned phenomena under the empathy umbrella---does not require a differentiation between self and other.  Further, many non-human empathy tasks---e.g. freezing when another rat freezes \citep{atsak2011experience}---do not require an appreciation of the differences between what oneself is feeling and a target is feeling, and may even manifest only because of this lack of appreciation.
\\

\emph{Motivation and action.}  Two points of significant divergence among empathy definitions are whether one needs to have a particular kind of other-oriented motivation to have empathy (row 13), or must take a particular kind of other-oriented action (row 14).  \citet{zaki2014empathy} emphasizes motivation, in general: ``empathy is often a motivated phenomenon in which observers are driven either to experience empathy or to avoid it'' \citep{zaki2014empathy}.  Definitions of empathy that incorporate notions of ``empathic concern'' usually go further and specify that an empathizer’s motivation must be other-oriented with an ultimate goal of ``relieving the valued other’s need'' \citep{batson2008}.  Some definitions also require that empathy lead to appropriate actions towards others, such as ``responding with the appropriate prosocial and helpful behaviour'' \citep{oliveira2011responding}.

By contrast, other views of empathy conceive of empathy more as a ``passive process of information gathering'' \citep{van2011empathy}.  These views are particularly prominent in evolutionary or neuroscience accounts of empathy, which generally postulate that more sophisticated and complex phenomena within the empathy umbrella grew out of, co-opt, or interact with automatic interpersonal perception and automatic mirroring.  Evidence used to support these views include the physiological reactions we have when seeing facial expressions of emotion, and the types of automatic copying babies do of their parents and animals do of other animals \citep{heyes2018empathy}.  Even if empathy can theoretically be derived through other more intentional or motivated mechanisms, these views usually see the passive aspects of empathy as critical for typical empathy-mediated behaviors. 
\\

\emph{Interactions with targets.}  A final set of distinctions that vary across empathy definitions pertain to empathy’s targets.  For example, according to \citet{bagheri2021reinforcement}, empathy involves ``reactive outcomes'' which ``aim to alter or enhance the target’s affective state'' \citep{bagheri2021reinforcement}.  Some therapeutic empathy accounts have similar emphases.  Consider McCarthy’s axiom that ``the sole judge of empathy is the receiver'' \citep{mccarthy1992empathy}.  According to these types of views, empathy may not be considered fully intact or functional if it doesn’t have the intended effect on the target (row 15). 

Another way that empathy can involve targets is through communication (row 16).  Some empathy accounts, particularly those from medicine or therapy, assert that ``Empathy is not empathy if it is not communicated'' \citep{mccarthy1992empathy}.  They may require that empathy include an essential ``action'' component during which a physician ``communicates understanding by checking back with the patient'' through words \citep{coulehan2001let}, ``tone of voice, facial expressions, body posture, and natural gestures'' \citep{guidi2021empathy}.  Similar requirements are emphasized in therapeutic contexts, where empathy is frequently treated as a mechanism for helping patients to feel understood and recognized, and therapists and significant others are encouraged to cultivate empathy by verbally acknowledging and validating what patients and partners think, feel, or are experiencing. 

A less obvious way empathy’s effects on the target could matter is if one’s theory of empathy or empathic learning requires emotional synchrony, motor synchrony, or temporal coordination between the empathizer and target.  For \citet{lim2015developing}, empathy involves ``synchronized reactions'' between the empathizer and the target \citep{lim2015developing, lim2015recipe}, and \citet{hatfield1993emotional} define primitive empathy as ``the tendency to automatically mimic and synchronize facial expressions, vocalizations, postures, and movements with those of another person’s and, consequently, to converge emotionally'' \citep{hatfield1993emotional}.  

Different still, interactions with targets could matter if and when empathy depends on integrating feedback from a target.  When targets do not convey what they are feeling or thinking clearly, empathizers may need to monitor the effects of ``their initial empathic response and adapt this response accordingly'' \citep{kozakevich2021adaptive} as more information becomes available about both the accuracy of what they predict about the target and the impact of the empathy they are communicating.  At least some of that feedback can’t be collected if the empathizer doesn’t communicate their thoughts or feelings related to the target successfully to the target.

The service industry literature employs a notion of empathy that incorporates yet another type of impact on the target.  In service settings, expressing empathy may be described as showing humility or vulnerability (row 17), with hopes of motivating users to be more forgiving of AI’s performance failures \citep{chi2023investigating}.  For example, one of the empathic responses \citet{lv2022artificial} programmed its AI system to have was ``Sorry, I know you must think I am stupid, but please give me a chance to make up for it'' \citep{lv2022artificial}.   The authors hypothesized that ``when AI gives an empathic response'' like this one, ``customers will follow social rules and interpersonal interaction practices, temporarily restraining their dissatisfaction, communicating with the AI in a more friendly manner, and collaborating to solve problems'', ultimately continuing to interact with the AI, even when it makes mistakes.  

Advice from popular psychology resources also allude to vulnerability on the part of the empathizer.  In a highly viewed video, Brené Brown said ``empathy is a choice, and it’s a vulnerable choice.''   Similarly, Helpguide.org reports ``Being empathetic requires you to make yourself vulnerable. When you hide behind an air of indifference, you make it harder for other people to trust or understand you. You also hold yourself back from feeling and understanding the full range of other people’s emotions.''  So vulnerability may be important for eliciting certain responses from targets, but also may be important for accurately assessing and comprehending the significance of what the target is going through.\\

\emph{Important take-aways.}  Acknowledging that the ``fine cuts'' of empathy and conflicts between published empathy theories can initially be overwhelming, we want to emphasize three important points.  First, awareness of the ``fine cuts'' of empathy can be empowering to the field of empathic AI, as we will discuss further below.  Second, we do not yet have sufficient empirical evidence to know which constellations of empathic abilities are necessary, helpful, or harmful in different contexts or when experienced by different individuals.  Third, even without this evidence, it seems likely that types of empathy that are useful in one context can be insufficient---or even counterproductive or unethical (as we will discuss)---in other contexts.  Next, we will analyze the empathic capabilities that would likely be needed by the medical empathic AI use cases described earlier to illustrate this point.

\subsection{}

\begin{table*}[t]
    \centering

\small
\begin{tabular}{|wc{0.4cm}|P{0.1cm}p{0.1cm}|P{1.4cm}|P{1.4cm}|P{1.4cm}|}
\hline
\textbf{Row} &
  \multicolumn{2}{c|}{\textbf{``Fine Cut'' of Empathy}} &
  
\scl{\textbf{Question}\\\textbf{Answerer}} &
  \scl{\textbf{Care}\\\textbf{Assistant}} &
  \scl{\textbf{Care}\\\textbf{Provider}} \\
  \hline
  
\textbf{1} &
  \multicolumn{1}{c|}{\multirow{3}{*}{What needs to be known about \textbf{T}?}} &   \textbf{P}erceptions &    &    \checkmark &  \checkmark
   \\ \cline{1-1} \cline{3-6} 
\textbf{2} & \multicolumn{1}{c|}{}                                  & \textbf{B}eliefs/Knowledge &  (\checkmark)     & \checkmark & \checkmark      \\ \cline{1-1} \cline{3-6} 
\textbf{3} & \multicolumn{1}{c|}{}                                  & \textbf{F}eelings          & (\checkmark)      & \checkmark   & \checkmark     \\ \hline
\textbf{4} & \multicolumn{1}{c|}{\multirow{3}{*} {How is it known by \textbf{E}?}} & \textbf{P}erceptions &     & \checkmark & \checkmark
   \\ \cline{1-1} \cline{3-6} 
\textbf{5} & \multicolumn{1}{c|}{}                                  & \textbf{B}eliefs/Knowledge & \checkmark & \checkmark  & \checkmark     \\ \cline{1-1} \cline{3-6} 
\textbf{6} & \multicolumn{1}{c|}{}                                  & \textbf{F}eelings          &      &   & \checkmark \\ \hline
\textbf{7} & \multicolumn{2}{l|}{\textbf{E} required to have \textbf{conscious} awareness of info?}                     &      &   &  \checkmark    \\ \hline
\textbf{8} & \multicolumn{2}{l|}{\textbf{E} required to make \textbf{accurate} theories or inferences?}                 & (\checkmark)     & \checkmark  & (\checkmark)      \\ \hline
\textbf{9} & \multicolumn{2}{l|}{\textbf{E} required to have \textbf{same} P/B/F as T?}                                 &  (\checkmark*)    & (\checkmark*)  &      \\ \hline
\textbf{10} & \multicolumn{2}{l|}{\textbf{E} required to have emotional response of a \textbf{certain valence}?}         & (\checkmark*)     & (\checkmark*)  & \checkmark     \\ \hline
\textbf{11} & \multicolumn{2}{l|}{\textbf{E} required to have \textbf{other-oriented emotional response}?}               & (\checkmark*)     & (\checkmark*)  & \checkmark     \\ \hline
\textbf{12} & \multicolumn{2}{l|}{\textbf{E} required to have \textbf{sense of the differences} between own experiences and those of T?}     & (\checkmark*)    & (\checkmark*)  & \checkmark     \\ \hline
\textbf{13} & \multicolumn{2}{l|}{\textbf{E} required to be appropriately \textbf{motivated}?}                           & (\checkmark*)     & (\checkmark*)  & \checkmark     \\ \hline
\textbf{14} & \multicolumn{2}{l|}{\textbf{E} required to take appropriate \textbf{action}?}                              & \checkmark   & \checkmark   &  \checkmark    \\ \hline
\textbf{15} & \multicolumn{2}{l|}{\textbf{E} required to have a \textbf{certain impact} on T?}                           & \checkmark      & \checkmark   &  \checkmark     \\ \hline
\textbf{16} & \multicolumn{2}{l|}{\textbf{E} required to \textbf{communicate} their feelings or knowledge successfully?} &   \checkmark    &  \checkmark  &  \checkmark    \\ \hline
\textbf{17} & \multicolumn{2}{l|}{\textbf{E} required to show some type of \textbf{vulnerability} to T?}                 &      &   &  (\checkmark)    \\ \hline

    \end{tabular}
     \caption{Empathy "fine cuts" needed for different empathic AI use cases. `E' = empathizer; `T'= target (i.e. person for whom E has empathy).  Parentheses indicates empathic capability could be useful in some cases, but is not necessarily required in all cases.  Asterisk indicates capability is mimicked.}
    \label{tab:my_label}
\end{table*}

\section{What Empathic Capabilities Do AIs Need?}

The empathic capabilities needed by question-answerers, care-assistants, and care-providers would likely have both similarities and differences.  All three AI systems, by definition, must take appropriate actions on behalf of their target (row 14; answering questions in a personalized way counts as such an action) and have certain impacts on their target (row 15) in order for consumers to continue using them.  To direct the appropriate actions, all the applications could benefit from accurate (row 8) knowledge about the target’s feelings (row 3) and beliefs (row 2), even if such knowledge is helpful without being necessary in the case of the question-answerer. 

In some scenarios it could be useful for the care-assistants and care providers to have knowledge of the target’s perceptions as well (row 1).  If an AI care-assistant needs to help a disoriented patient find their way home safely while walking around the block, for example, the AI would benefit from knowing whether the target is noticing stop signs, traffic lights, and traffic, and whether the target can hear the AI’s instructions.   Overall, the more any medical AI knows and appropriately addresses specific concerns or thoughts targets have, the more useful the AI is likely to be (as long as doing so doesn’t creep users out).  It is plausible that effective AI care-assistants or care-providers could be trained using black-box algorithms that call into question whether the AI really has explicit ``knowledge'' of targets’ feelings, beliefs, or perceptions, but we will put that issue aside for now.  

It seems sufficient for AI question-answerers and care assistants to obtain knowledge about the targets’ beliefs, feelings, and possibly perceptions solely through beliefs (in the form of statistical models) about the world (row 5), but many (if not all) AI care-assistants may also need to obtain this information through their own artificial perceptions (row 4), such as through video feeds or sensor readings of the target.  According to the criteria established earlier, AI care-takers would have to meet an even higher standard, though.  The expectation of care-takers as they have been described previously and as we are conceiving of them here is that they need to glean knowledge about the target through their own “real and appropriate feelings” that are shared, at least to some degree, with the target (row 6).   

The depth and degree of accuracy (row 8) required by each AI application could plausibly differ.  It could be sufficient for AI question-answerers to correctly predict the valence of the question-asker’s feelings or beliefs in order to craft acceptable responses for that situation, rather than correctly predict exactly what the question-asker is thinking or feeling.  Care-assistants, on the other hand, could need to have more detailed and accurate knowledge of a target’s beliefs and feelings to help the target with challenges like finding their way home, agreeing to take medications they are reluctant to take, or providing entertainment.  AI care-takers would have similar accuracy requirements, but if they are perceived to genuinely care about the target and be motivated to help them, could be given more leeway to sometimes be incorrect.

In all three medical AI applications, the AIs need to be able to communicate their model (or understanding) of the target’s situation to the target (row 16), at least to some degree.  For instance, part of the reason ChatGPT’s answers to medical questions asked in Reddit’s r/AskDocs forum came across as more empathic than answers given by human physicians is that ChatGPT made statements like ``It’s natural to be concerned...'' \citep{ayers2023comparing, kolata2023doctors}.  These statements do not attempt to mimic feelings, but they acknowledge what the target is likely experiencing and normalize those experiences.  It seems likely that people also need to feel that an AI care-assistant or caretaker has sufficient knowledge of their situation, experiences, and feelings for their guidance to be trustworthy.  This trust will only be gained if the AIs can state what they believe to be true about the targets in approachable ways, so that the target can determine whether the AI’s assessment is correct enough to warrant following its advice that is supposed to be given on the target’s behalf.  It is possible that AIs may benefit from showing vulnerability or humility when they communicate with targets (row 17), but there’s no obvious reason to assume it is necessary for any of them to do so, although some have argued vulnerability is needed to show care.  

The requirements for the different AI applications differ more in rows 7 through 13 of Table 1.  It seems sufficient for the AI question-answerer and care-assistant to have a model of what the target feels, believes, and in some cases, perceives.  They do not need to be conscious (row 7), have the same feelings, beliefs, or perspectives as the target (row 9), or have a sense of the differences between their own experiences and the target’s experiences (row 12) to achieve their purposes for their users.  That said, it is possible that they would benefit from mimicking (indicated with an asterisk in Table 1) the responses (through words, facial, or body actions) of an entity that has the same feelings as the target (row 9), that has feelings of an appropriate valence (row 10), that is other-oriented (row  11), that differentiates between their experiences and those of the target (row 12), or that is appropriately motivated (row 13) \citep{montemayor2022principle, shamay2022inter, perry2013understanding, jeffrey2016empathy}.  For example, users responded favorably when ChatGPT responded ``I’m sorry to hear that you got bleach splashed in your eye'', even though ChatGPT has no way to feel sorry \citep{ayers2023comparing}.  

Users of AI care-assistants may also find comfort in an AI that gives responses like, ``It makes me really angry that your boss treated you that way.  I really care about you, and am here to support you'' or ``I know it might be irritating to hear my medication reminders so frequently.  I just really want to make sure you stay healthy!'', even if the AI has no ability to feel angry, care about anything, or want something.  On the other hand, it is also possible that human users might find AIs that pretend to have these types of capabilities annoying, eerie, untrustworthy, or unethical \citep{shao2023empathetic, seitz2024artificial}.  More research needs to be done about human preferences for AIs with these qualities, but AI question-answerers and care-assistants do not need to have subjective experiences in order to make progress towards their intended purposes.

The AI caretaker, in contrast, would need to have real subjective feelings to achieve its intended impact, at least according to criteria described.  Even if future research finds such feelings are not needed for all people to feel cared for in all circumstances, it is sufficient for our current purposes to focus on situations where they are necessary.  In these situations, the AI’s feelings would need to be of the appropriate valence (positive when something good happens to the target and negative when something bad happens to the target; row 10), target-focused (based on the target’s welfare rather than the AI’s welfare; row 11), and causally linked to actions the AI reliably takes on the target’s behalf (row 14).  Further, it is plausible that the people being cared for want their caretakers to experience their feelings consciously (row 7), according to prevalent views of what is needed for targets to feel cared for without being misled.  It’s not clear that the AI caretaker would ever need to feel the exact same feelings as the target (row 9), even if doing so can be a mechanism for target-focused emotions and actions in humans.  AI caretaker’s should be able to differentiate their experiences from the target’s experiences (row 12), though, and be motivated primarily by the target’s experience (row 13).  

We focused our discussion on three prevalent empathic use cases in medicine, but there are numerous empathic use cases in other domains as well, and they may require their own constellations of empathic capabilities.  Some empathic AIs in transportation, for example, are developed to allow an autonomous vehicle to navigate interactions with other cars safely by taking into account other drivers’ predicted actions and incentives.  A set of researchers working on this application described the empathy they were interested in as ``the ability to understand others' intent by simultaneously inferring others' understanding of the agent's self intent'' \citep{ren2019shall}.  A different set of researchers working on this application described a theoretical framework for an ``empathic'' autonomous agent that they said ``proactively identifies potential conflicts of interests in interactions with other agents (and humans) by considering their utility functions and comparing them with its own preferences using a system of shared values to find a solution all agents consider acceptable'' \citep{kampik2019empathic}.  Unlike AI medical question answerers, these empathic AI ``social negotiation managers'' need to know at least a critical subset of the target’s beliefs and preferences, such as (but not limited to) what targets think the risks associated with different possible actions are likely to be, what they assume other agents in the situation will do, what risks they think are acceptable to take, and what they intend to do.  It also seems fairly important for the social negotiation manager to know what its targets perceive in their environment, such as whether it is possible for them to see all the cars and blinker signals in nearby lanes on a highway or all the pedestrians trying to cross the street at an intersection.  Unlike AI care-assistants or care-providers, such social negotiation managers only need to know the target’s emotions insomuch as they are likely to affect the target’s actions (e.g. drivers that are angry drive faster and are more likely to engage in behaviors that endanger others).  Many other differences would exist in the rows of Table 1 as well, as would be the case for empathic AIs developed for other domains like marketing, gaming, education, or virtual friendship.

\section{Implications for AI Creators and Users} 

Table 1 and the discussion in the last section should make it clear that ''Empathic AI'' need not, and likely should not, refer to one thing with a singular set of abilities.  Instead, we posit that it will be more fruitful for ''Empathic AI'' to be used as a term that refers to agents with different combinations of empathy-related capabilities selected to address the needs of a particular use case.  Rather than try to build a system that matches a specific empathy definition, we recommend that AI creators commit to thinking deeply about which of the components discussed earlier are required, optional, or incompatible with the type of empathy needed for their chosen use case, and collect evidence when necessary to support those conclusions.  Doing so will help both empathic AI creators and users in many ways.  
\linebreak

\emph{Knowing what to build.}  Some groups are trying to create general AI that can teach itself how to do anything it needs to do, and may hope that empathy is one of the things AIs learn or develop along the way.  Other empathic AI systems are examples of narrow AI that are used to tackle individual tasks close to those their creators explicitly and intentionally designed them to achieve.  Within narrow AI systems, it is common for creators to have to design different approaches for managing each distinct subtask or function the AI needs to perform, and then combine those approaches into one integrated system.  Given that each approach requires significant investment, the systems that are ultimately built are less likely to have specific empathic capabilities or constraints if system creators don’t intentionally try to build them in.

As an example, many artificial empathy systems that have been designed to date analyze real-time video feeds of a human target, try to identify what emotion the human target is feeling based on their facial expressions, and then implement a response the AI has learned is most appropriate for a target’s identified emotion (e.g. \citealt{xiao2016computational}).  These systems, as currently implemented, typically have no ability to predict what a human target sees or knows from their own sensory input.  Therefore, these AI systems currently have no way of representing the critical perceptual information that is needed to pass the Sally-Ann test described earlier.  They also have no way of representing or understanding the thoughts the target is having that cause the emotion the target is expressing.  That means these empathic AI systems would not be considered empathic by accounts that require the target to have beliefs about the target’s perceptions and beliefs, and wouldn’t meet the requirements of the AI care assistants we discussed earlier.  Nonetheless, it is likely technically feasible for the systems to be adapted so they could do so.  
There is nothing stopping AI creators from making these adaptations in principle, other than that they may not have realized these criteria are important for some definitions and applications of empathy that they have not yet had reason to work on.
\linebreak 

\emph{Avoid wasting time on unnecessary features.}  AI creators may turn to the empathy literature to design requirements for an empathic AI system, and unknowingly choose an empathy definition that is not well-suited to their use case, leading them to waste time and resources building unnecessary features into their system.  For instance, although mirroring is likely an important mechanism of some types of human empathy and its evolutionary precursors, AIs do not need to use the same mirroring mechanisms to be able to accurately predict what a target is thinking, feeling, or perceiving, or to respond to those thoughts, feelings, or perceptions appropriately.  AI designers can of course incorporate mirroring mechanisms into their empathic AI system designs if they wish or find it useful.  Other designs may be more technically feasible or efficient, though, and there is no obvious reason why all artificial systems should be required to ``match'' their targets to be considered empathic when some applications would not benefit from such matching.  

As another example, the fact that we do not yet know whether AI will ever have consciousness does not in principle preclude AI from having at least some types of empathy.  Although some conceptions of human empathy require humans to be consciously aware of what targets are feeling, AI that lacks consciousness but meets requirements of other conceptions of empathy might still be extremely useful in appropriate situations.  The situations addressed by AI medical question-answerers and care assistants we discussed earlier are examples, and there are many more.  
\linebreak 

\emph{Facilitating ethical disclosure and use.} If users think an AI has empathy according to one definition but the AI really only has empathy according to a different definition, users can be misled, exploited, make poor choices, or even be caused unnecessary emotional harm.  In doing so, users would not be treated with the respect many agree is due to all humans, including (if not especially) vulnerable humans \citep{korsgaard2021valuing}.  

Consider observations that patients are more likely to take their prescribed medication, follow recommended lifestyle changes, share diagnosis-relevant personal information, and experience relief from mental health challenges when they feel their doctors or therapists care for them \citep{kim2004effects, crockett2010serotonin, eide2004listening, finset2017empathy}.  Multiple entities are busy trying to make medical AI care assistants and care providers that humans perceive to be sufficiently empathic to lead to similar positive health outcomes.  As we have discussed, it is frequently argued that empathy is as impactful as it is in these contexts because it makes targets feel their welfare genuinely matters to somebody else in a deep and personal way.  However, even the most ambitious empathic AI systems that are available today will not have the types of feelings and social motivations required to care about a human in the way humans care about each other.  In other words, we currently don’t know how to create these types of AI care providers, and we don’t know if it will ever be possible to create them.  Yet, AI care assistant systems are being built to have the appearance of having those feelings and motivations, and some users incorrectly interpret the AIs' behavior as evidence the AI cares about them and has genuine feelings \citep{brandtzaeg2022my, hu2023social}.  Empathic AI creators must make clear to users what abilities their AI systems do and do not have in these situations to minimize user deception and exploitation.  One approach for doing so would be to leverage adapted model cards or "nutrition labels" that disclose which of the empathy "fine cuts" we described here are present or absent in a specific system \citep{mitchell2019model}.  

It is also critical to appreciate that additional ethical problems will arise, even if AI creators do their best to inform users of their AIs’ abilities and lack of abilities.  Some vulnerable users, such as elderly patients with dementia, may not understand that an AI they are interacting with is not a living system, and may choose to make sacrifices on the AI’s behalf the same way they might for another human they had a mutually caring relationship with.  Others might know and lament that an AI isn’t able to care for them the same way a human can, but voluntarily engage with the AI nonetheless because they feel they have no better alternative.  Such interactions can certainly have benefits \citep{inzlicht2023praise}, but they could also end up making patients feel even more alone and even more undeserving of human understanding, compassion, and love \citep{montemayor2022principle, perry2023ai}.  Ubiquity of AI companions could also erode society’s commitment to believing all people deserve human versions of these things more generally.  

Different ethical concerns arise when targets incorrectly assume an empathic AI’s empathy—even if it is unconscious and unfeeling—is designed to function on behalf of the target (rows 11, 13, 14, and 15 in Table 1). Some empathic AIs will be designed this way.  Others, however, could qualify as empathic by some definitions, even though they are designed to function primarily on behalf of the AI creators. Empathic AI of this kind is often discussed in customer service contexts, such as Siena AI’s chatbot that is advertised to have ``human empathy in every interaction''.  Targets can benefit from the AI's empathy through having more positive customer experiences, but the point of the empathy in such cases is still primarily to improve metrics the AI creators care about (such as customer retention or conversion).  Sometimes what benefits an AI’s target and the AI’s creators will conflict; AIs that leverage their empathy to create highly personalized experiences for targets may be enjoyable, but may also be used to more effectively nudge or motivate users to do things that are not to their benefit, like buy harmful products, gamble when they cannot afford to do so, or share private information.  Users will be more susceptible to empathic AI-driven manipulation if they fall prey to the assumption that all kinds of empathy are good for the target.  It’s plausible that being transparent about what kind of empathy is being implemented in a given AI will mitigate this ethical risk, but determining what kinds and forms of transparency are sufficient, and how these might differ across various user segments, are open and important questions. 

We will not attempt to settle these issues here.  Our point is that creators of empathic AI will not be able to anticipate or analyze these kinds of ethical concerns adequately if they do not appreciate the distinctions in Figure 1 sufficiently. 
\\

\emph{Highlighting potential divergences between artificial and human empathy needs.}  The overall endeavor of considering what capabilities empathic AIs in different contexts need also brings into focus differences between the requirements of human and artificial empathy.  One that we have already alluded to is that artificial empathy may need to rely on accurate perception of the target’s feelings, beliefs, and perceptions more than many conceptions of human empathy.  Some empathy accounts want it to be possible for an empathizer to have empathy solely through imagining—but not perceiving—something happening to another, or even through thinking about the potential of something happening to another without fully imagining what that something would be like.  There may be some situations where artificial empathy that functions without mechanisms for perceiving what a target is perceiving or feeling is sufficient for its intended purpose, like the medical question-answerer we discussed earlier.  In many other potential artificial empathy applications, though, having accurate empathic perceptions will be absolutely essential and must be a critical part of the system’s design, even if traditional empathy theory does not always emphasize this capability.

We also alluded to the possibility that the accuracy of artificial empathic perception could be held to a higher standard than human empathic perception in order to be considered successful in serving its intended purpose.  Humans who are incorrect about what they perceive, think, or feel a target is going through are often still considered to be empathic as long as they are believed to be genuinely trying to understand a target or act on the target’s behalf.  Empathic AIs that are not conscious and not believed to be motivated in the same way as humans cannot fall back on this mechanism for having empathy ascribed to them, and may therefore need to label what targets perceive, think, or feel a target is going through accurately (and maybe very accurately) in order to be considered empathic.  Some may even equate the level of empathy artificial agents have with the accuracy of their labels of targets’ experiences.  In addition to the way this standard diverges from human empathy, it is notable because identifying human emotions from images or videos using AI still remains a formidable challenge \citep{yan2021emotion}, facial expressions or verbalizations do not always reflect a person's feelings \citep{barrett2019emotional}, and there are serious debates about whether and how human emotions should be understood, distinguished, measured, and described anyway \citep{barrett2021navigating}.  
\\

\emph{Highlighting research needs.}  Analyzing what capabilities empathic AIs need in different contexts also highlights important open questions and areas of research that are needed to move the interdisciplinary empathic AI field forward.  One of the most notable is that the empathy research community has surprisingly little evidence or even theory about what aspects of the capabilities under the empathy umbrella might be important for a given application, and we have even less information about what kind of empathy users want their AI systems to have.  For example, many assume users want their AI assistants to express sympathy and encouragement during interactions.  However, it is possible that users find that kind of expression from an AI annoyingly disingenuous and instead just want AI systems to anticipate users’ desires and knowledge well enough to make effective suggestions or make good decisions on the users’ behalf \citep{seitz2024artificial}.  Likewise, some have proposed that building embodied versions of empathy into AI systems might make them more likely to comprehend and comply with social and moral norms \citep{bennett2021philosophical}.  Even if that ends up being true, such embodied capabilities have many similarities to human feelings, and users may not think it is ethical to build AI systems with those abilities.  In sum, builders of empathic AI need to think not only about what computations their AIs need to be able to achieve, but also how humans will receive and interact with AI systems that perform—or appear to perform—those computations.  Empathy researchers and ethicists often assume the answers to questions about what kind of empathy is desirable are self-evident, but they are not, and research is needed to address these questions empirically.  

A second area for development is methods that can determine whether an empathic AI system has the specific capabilities under the empathy umbrella it is designed to have.  Many empathy assessments used in humans rely on self-reports of what one feels or how one responds to a target, which can be unreliable and biased by efforts to respond in a socially favorable way.  Not only is it unclear what it means for an AI to ``self-report'' using such assessments, it is likely that some AIs would respond with hallucinatory and misleading answers.  

Further, current empathy assessments are plagued by many mismatches between what abilities the assessments draw on and the empathy definitions the assessments are meant to index \citep{hall2022empathy}.  Sometimes the assessments require more abilities than the chosen empathy definitions do.  As we have highlighted, empathy definitions do not always explicitly require that empathizers have certain perceptions, and may even emphasize that empathizers should be able to imagine and have appropriate reactions to what a target thinks or feels without seeing or hearing the target’s reactions.  It is often underappreciated, though, that many tasks meant to assess the presence or absence of empathy according to those same definitions require accurate perceptions on the part of the empathizer.  This is important because, by some accounts, agents can still have intact empathy abilities even if they are unable to accurately perceive or interpret the things required to pass empathy tests.  Researchers have argued this phenomenon explains why people with autism have inappropriately been labeled as lacking empathy; people with autism may have challenges attending to and interpreting others’ emotional cues, but once those cues are identified and correctly labeled, have strong empathic responses (that are sometimes even more intense than those experienced by neurotypicals; \citealt{fletcher2020autism}).  Similarly, an AI might have some abilities within the empathy umbrella but not be able to pass empathy tests that require additional abilities the AI does not yet have, even if those abilities are not specific to empathy.

The inverse also occurs: some empathy tasks can be completed successfully without the use of capabilities or phenomena included in a given definition of empathy.  Interpretations of rodent empathy tasks, in particular, must manage this challenge.  One such rodent empathy task tests how much ``empathizing'' rodents freeze—a behavior historically associated with fear—when they see another rodent freeze \citep{atsak2011experience}.  Another type of rodent empathy task measures pain-related writhing when an animal is next to another animal who is also exhibiting pain-related writhing \citep{langford2006social}.  Others involve releasing a target rodent trapped inside a Plexiglas tube or opening a door so that a target rodent in a pool of water can access a dry platform (Mason, 2021).  

All of these rodent tasks have been described as models for empathy, primitive forms of empathy \citep{keum2016rodent}, or ``empathically-driven targeted helping behaviors'' \citep{cox2021current}. However, since it is very hard to know what an animal is subjectively feeling when they perform empathy tasks (despite laudable efforts to discern those feelings through clever experimental designs), we can’t be sure whether rodent ``empathizers'' meet the criteria in many of the rows in Table 1.  AI researchers trying to develop empathic AI are likely to run into similar problems when trying to establish methods for testing whether their AIs have “empathy”, irrespective of the substantial technical progress being made towards developing some individual artificial empathic capabilities listed in Table 1  \citep{khare2023emotion, raamkumar2022empathetic, shapira2023clever}.  

For all these reasons, investment is needed to develop robust assessments for empathic AI systems, and doing so will be a considerable challenge.  In the meantime, AI creators and researchers across fields must take care to distinguish which empathic capabilities empathy tasks do and do not require to pass, and be aware that their tests may rely on capabilities (like perceptual abilities) outside of some empathy definitions.

A third area for development is empathy theories, definitions, and models that can better account for the role targets’ perceptions of empathy and the interactions between empathizers and targets play in how empathy works.  Empathic AI products will only be successful if they achieve certain intended impacts on their targets.  Most psychology and neuroscience accounts and tests of empathy, though, focus primarily on what the empathizer perceives, knows, feels, or experiences.  Some empathy accounts do incorporate notions of how empathizers and targets interact, or even focus on such interactions (Rijnders, Terburg, Bos, Kempes, \& van Honk, 2021; Zaki, Bolger, \& Ochsner, 2008).  Shamay-Tsoory has recently emphasized the role empathy has on a target’s distress \citep{shamay2022inter}. \citet{main2017interpersonal} wrote ``in real life empathy is an interactive social process dependent upon both individuals for adaptive functioning'' \citep{main2017interpersonal}, and \citet{kozakevich2021adaptive} introduced the idea of ``adaptive empathy'' that is responsive to feedback \citep{kozakevich2021adaptive}.   There are also threads of empathy research that implicitly recognize interactions between empathizers and targets.  Most notably, it has been shown that our facial and body movements, physiological responses, and even brain responses synchronize over time with targets when we empathize or feel empathized with \citep{levy2019synchronous}, and disrupting this synchrony can interfere with the empathy felt, expressed, or perceived \citep{koehne2016perceived}.  Still, many popular empathy accounts, syntheses, and meta-analyses of empathy research ignore this interactive aspect of empathy, and there is still relatively little theory or data about how and when targets feel or perceive different types of empathy, or how targets’ responses to perceived empathy impact the empathy an empathizer in turn experiences.  More work is needed in this area for AI empathic products to have their intended impact, and to be able to analyze the potential benefits and harms of an empathic AI accurately.  

\section{Conclusion}

Artificial empathy has many exciting uses and has the potential to overcome many of the disadvantages of human empathy by virtue of being free of the computational constraints and ingroup biases of human empathy \citep{inzlicht2023praise}.  It also has the potential to harm individuals and society more generally if developed unthoughtfully and without sufficient appreciation for the functions of its different characteristics, like if an empathic AI encourages a user to follow through with their proposed suicide attempts in an misguided effort to validate the user's feelings.  To realize the possible advantages of empathic AI while successfully navigating its ethical challenges, we first must be clear about what AIs need to do or be able to do to be considered empathic.  We have argued that different constellations of phenomena within the empathy umbrella will be needed and desired for different empathic AI applications, and that the criteria used to assess each phenomenon may be different for AIs and humans functioning in the same context.  More evidence needs to be collected to determine what principles best govern what constellations of empathic phenomena are either necessary or desirable in what situations, and more precise and robust assessments need to be developed to test the presence or absence of those empathic phenomena.  As empathic AI applications are pursued, both AI creators and empathy researchers must think deeply about the contingencies between an empathizer and target.  All of these considerations and evidence should be incorporated into ongoing discussions about empathic AI's ethical implications.  Our aim here has been to amplify and focus future work in this area---work that is poised to impact society at an increasingly large scale.

\section*{Acknowledgements}

We thank Walter Sinnott-Armstrong, Daryl Cameron, Vijay Keswani, and members of the Duke Moral Attitudes and Decisions Lab for their comments on initial drafts of this manuscript.  Jana Schaich Borg's work is supported in part by a research grant from OpenAI.

\bibliography{references}


\end{document}